\pdfoutput=1
\documentclass{article}
\usepackage{arxiv}

\usepackage[utf8]{inputenc}
\usepackage[T1]{fontenc}
\usepackage{amsmath,amssymb,amsfonts,bm}
\usepackage{graphicx}
\usepackage{booktabs}
\usepackage{multirow}
\usepackage{array}
\usepackage[dvipsnames]{xcolor}
\usepackage{microtype}
\usepackage{enumitem}
\usepackage{wrapfig}
\usepackage[font=small,labelfont=bf,skip=4pt]{caption}
\usepackage{subcaption}
\usepackage{xspace}
\usepackage[round,semicolon]{natbib}
\usepackage{url}
\definecolor{linkblue}{RGB}{0,72,150}
\usepackage[colorlinks=true,linkcolor=linkblue,citecolor=linkblue,urlcolor=linkblue]{hyperref}

\newcommand{\method}{\textsc{DynamicHOI}\xspace}
\newcommand{\repro}{\ensuremath{^{\dagger}}}
\newcommand{\tablesize}{\fontsize{8}{9.5}\selectfont}

\title{DynamicHOI: Coupled Dynamics for\\Physics-aware HOI Reconstruction}
\renewcommand{\shorttitle}{\textsc{DynamicHOI}: Coupled Dynamics for Physics-aware HOI Reconstruction}
\date{}

\author{Wenliang Guo \qquad Zhanbo Huang \qquad Yu Kong\\
Michigan State University}

\hypersetup{
  pdftitle={DynamicHOI: Coupled Dynamics for Physics-aware HOI Reconstruction},
  pdfauthor={Wenliang Guo, Zhanbo Huang, Yu Kong},
}

\begin{document}

\maketitle

\begin{abstract}
We study hand-object interaction (HOI) reconstruction from monocular RGB videos, where partial observations can produce visually plausible yet mechanically inconsistent trajectories. Existing methods mainly enforce visual and geometric agreement, leaving the underlying interaction dynamics insufficiently constrained. We propose \method, a physics-aware HOI reconstruction framework combining geometry-grounded diffusion refinement with coupled hand-object dynamics. Geometry spatially grounds visual evidence for trajectory refinement, while articulated inverse dynamics and Newton-Euler dynamics derive hand generalized forces and object wrenches for dynamics-level supervision. We further couple hand and object dynamics through contact-force transfer and recover active hand actuation as an interaction-level physical quantity. We formulate its empirical magnitude distribution into a probabilistic prior that penalizes unlikely actuation and suppresses mechanically implausible reconstructed motion. Experiments on three HOI datasets show consistent improvements in both hand and object reconstruction. The reconstructed trajectories further benefit downstream applications including hand world-model generation and robotic manipulation learning, demonstrating the value of physics-aware HOI modeling beyond reconstruction. Project page: \url{https://wenliangguo.github.io/HOI-Reconstruction-Page/}.
\end{abstract}

\section{Introduction}
\label{sec:intro}

Hands are the primary interface through which humans perform everyday hand-object interactions (HOIs). Reconstructing these interactions from monocular videos into temporal 3D hand-object mesh trajectories is essential for immersive technologies \citep{zhou2023mixed,feng2020resolving}, embodied learning \citep{qin2022dexmv,chen2025vividex}, and world modeling \citep{li2026egohoi}. Monocular videos provide only partial observations under occlusion and viewpoint ambiguity, allowing multiple 3D trajectories to explain similar visual evidence and making reliable reconstruction challenging.

Recent work has substantially advanced HOI reconstruction. Optimization-based methods jointly fit hand, object, and scene representations \citep{huang2022hhor,ye2023diffhoi,fan2024hold}, while feed-forward approaches exploit contact, occlusion, and learned interaction priors \citep{chen2022alignsdf,wang2025magichoi,aboukhadra2026ghost,wang2026arthoi}. Recent methods have also extended HOI reconstruction to egocentric videos \citep{zhang2025hawor,fu2026egograsp,ye2026whole}. Despite this progress, two challenges remain. First, geometry is not fully exploited to ground visual representations, limiting ambiguity resolution under occlusion and restricted viewpoints \citep{ye2026whole}. More importantly, reconstruction is typically supervised at the trajectory level, which encourages geometric accuracy but does not explicitly constrain the dynamics implied by the estimated motion. For example, a reconstructed object may suddenly translate or rotate during manipulation without a corresponding force being exerted by the hand. Dynamics provides complementary supervision by constraining the forces and torques required to produce reconstructed motion.

Introducing such dynamical constraints remains difficult because physical quantities such as forces are not directly observable from videos. \citet{zhang2024physpt} and \citet{ismayilzada2026padhand} pioneered articulated hand dynamics by deriving generalized forces from reconstructed trajectories through inverse dynamics, providing physical supervision beyond geometric states. However, these formulations characterize only the hand and do not account for the dynamics of the manipulated object. In HOI, object motion is governed by its own rigid-body dynamics, while the hand and object are mechanically coupled through contact: active hand actuation produces hand motion and generates interaction forces that drive the object. Therefore, modeling hand dynamics alone or treating the two bodies independently leaves this mechanical dependency unconstrained. Physics-aware HOI reconstruction thus requires both hand and object dynamics as well as their physical coupling through contact to form a unified dynamically-constrained system.

To address these challenges, we propose \method, a physics-aware HOI reconstruction framework that integrates geometry-grounded visual refinement with coupled hand-object dynamics. To better resolve visual ambiguity, we encode the hand-object geometry and use it to spatially ground visual evidence within a diffusion-based iterative refinement, providing geometry-aligned guidance for trajectory estimation. To further constrain the motion itself, we derive hand generalized forces through articulated inverse dynamics \citep{zhang2025diffusion,ismayilzada2026padhand} and rigid-object wrenches through Newton-Euler dynamics \citep{luh1980line,khalil2011dynamic,hu2022physicalinteraction,lutter2023differentiable}. Aligning these motion-implied physical quantities between reconstructed and ground-truth trajectories introduces dynamics-level supervision beyond trajectory alignment.

We further couple the hand and object through contact-force transfer \citep{hu2022physicalinteraction,nakajima2022prediction}. Active hand actuation jointly accounts for the generalized force required by hand motion and the contact reaction induced by driving the object, providing a shared physical explanation for both trajectories. Our key insight is that active actuation during everyday manipulation follows a characteristic statistical distribution \citep{tanghe2019probabilistic}. We estimate this empirical distribution and impose its negative log-likelihood as a probabilistic regularizer, penalizing reconstructions that induce low-probability actuation. Together, the hand dynamics, object dynamics, and actuation prior form a coupled dynamical system that provides physically grounded supervision for HOI reconstruction.

Experiments across three HOI reconstruction datasets \citep{banerjee2025hot3d,chao2021dexycb,hampali2020honnotate} demonstrate that \method consistently improves both hand and object reconstruction. We further use the reconstructed trajectories to condition hand world-model video generation \citep{li2026egohoi} and train dexterous robotic manipulation \citep{chen2025vividex}. Improvements in both applications show that physics-aware reconstruction provides more useful motion signals beyond reconstruction accuracy. In summary, our contributions are threefold:
\begin{itemize}[leftmargin=1.5em,itemsep=2pt,topsep=2pt]
    \item We present a geometry-grounded HOI reconstruction framework that uses 3D geometry to spatially ground visual evidence during diffusion refinement for more accurate trajectory estimation.
    \item We introduce dynamics-level supervision by aligning hand generalized forces and object wrenches between reconstructed and ground-truth trajectories to encourage physically consistent motion.
    \item We couple hand-object dynamics through contact-force transfer and regularize active hand actuation with a probabilistic prior to suppress reconstructions that induce implausible actuation.
\end{itemize}

\section{Related Work}
\label{sec:related}

\textbf{Hand and object reconstruction.}
Image-based hand models such as HaMeR \citep{pavlakos2024hamer} and WiLoR \citep{potamias2024wilor} learn strong visual priors for mesh regression, while video-based methods such as HaWoR \citep{zhang2025hawor} and Dyn-HaMR \citep{yu2025dynhamr} further exploit temporal context and camera motion to recover global hand trajectories. For objects, FoundationPose \citep{wen2024foundationpose} conditions 6D pose estimation on object geometry, whereas ForeHOI \citep{chen2026forehoi} combines visual evidence with 3D shape priors to reconstruct objects under hand occlusion. However, these methods do not explicitly leverage joint 3D hand-object geometry to organize visual evidence. In contrast, our framework grounds visual cues in the 3D HOI geometry within diffusion refinement, improving reconstruction under occlusion and monocular ambiguity.

\textbf{Hand-object interaction reconstruction.}
Joint HOI reconstruction reduces monocular ambiguity by exploiting dependencies between the hand and object. Early methods jointly estimate or refine their geometry and poses using contact and non-penetration constraints \citep{hasson2019joint,grady2021contactopt,chen2022alignsdf}. Video-based approaches further incorporate differentiable rendering, implicit representations, and learned interaction priors \citep{ye2023diffhoi,huang2022hhor,fan2024hold}, while recent methods improve generalization to limited viewpoints, unseen categories, articulated objects, and egocentric settings \citep{wang2025magichoi,aboukhadra2026ghost,wang2026arthoi,fu2026egograsp,ye2026whole}. These methods primarily enforce visual and geometric consistency of the motion, while our proposed framework focuses on modeling articulated-hand and rigid-object dynamics to constrain the physical requirements underlying their joint trajectories.

\textbf{Physics-aware HOI modeling.}
Physics-aware methods improve mechanical plausibility through explicit dynamics or learned physical constraints. Physical Interaction estimates contact forces that satisfy rigid-object dynamics from RGB-D observations \citep{hu2022physicalinteraction}, while Physics-Aware HOI Denoising refines hand motion given an accurate object trajectory \citep{luo2024physicsawarehoi}. Other hand-centric approaches introduce phase-dependent physical constraints \citep{zhang2025diffusion} or probabilistic articulated dynamics \citep{ismayilzada2026padhand}. Nevertheless, existing methods do not jointly model hand dynamics, object dynamics, and their coupling through contact. We address this gap by coupling the two dynamical systems through contact forces and regularizing the resulting active hand actuation with a probabilistic prior.

\section{Methodology}
\label{sec:method}

\begin{figure}[t]
    \centering
    \includegraphics[width=0.5\linewidth]{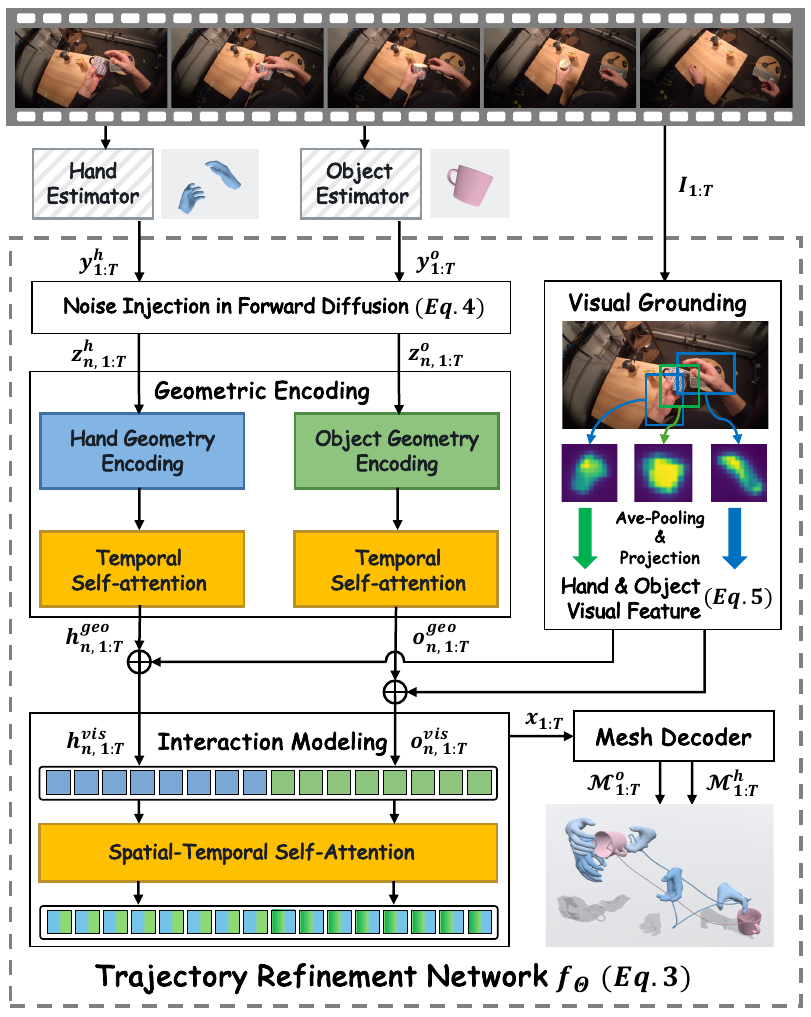}
  \caption{Framework overview. Framewise reconstruction models first estimate the initial trajectory in parametric states, which is then refined progressively by a diffusion-based refinement network, and finally decoded to HOI mesh trajectory.}
  \label{fig:method}
\end{figure}

\subsection{Problem Formulation}
\label{sec:formulation}
Given an RGB video $I_{1:T}=\{I_t\}_{t=1}^{T}$ and the canonical mesh $\mathcal M^o$ of the interacting object\footnote{Canonical mesh is the object 3D shape in a reference coordinate frame before any rotation or translation.}, HOI reconstruction aims to recover the 3D hand and object mesh trajectories over $T$ frames. Following prior work \citep{ismayilzada2026padhand,ye2026whole}, we represent the reconstruction using compact parametric states and then deterministically decode them into meshes. Specifically, a reconstruction model $\mathcal G_{\Theta}$ with learnable parameters $\Theta$ predicts the joint hand-object trajectory during inference
\begin{equation}
\mathbf x_{1:T}
=
\left\{
(\mathbf x_t^h,\mathbf x_t^o)
\right\}_{t=1}^{T}
=
\mathcal G_{\Theta}
\left(
I_{1:T},
\mathcal M^o
\right),
\label{eq:hoi_reconstruction_task}
\end{equation}
where $\mathbf x_t^h$ and $\mathbf x_t^o$ denote the hand and object parametric states at frame $t$, respectively. The hand state $\mathbf x_t^h=(\boldsymbol\rho_t^h,\mathbf t_t^h,\boldsymbol\beta^h)$ is represented by MANO parameters~\citep{romero2017mano}, where $\boldsymbol\rho_t^h\in\mathbb R^{16\times6}$ contains the 6D rotations of the global hand orientation and the $15$ articulated MANO joints, $\mathbf t_t^h\in\mathbb R^3$ denotes the global translation, and $\boldsymbol\beta^h\in\mathbb R^{10}$ represents the hand shape and remains fixed throughout the video. The estimated states $\mathbf x_{1:T}^h$ are decoded through the MANO layer \citep{romero2017mano} to obtain the hand mesh trajectory $\mathcal M_{1:T}^h$. The object state $\mathbf x_t^o=(\mathbf R_t^o,\mathbf t_t^o)$ is represented by its rigid pose, where $\mathbf R_t^o\in SO(3)$ and $\mathbf t_t^o\in\mathbb R^3$ denote its rotation and translation, respectively. The object mesh at frame $t$ is obtained through the rigid transformation
$\mathcal M_t^o=\mathbf R_t^o\mathcal M^o+\mathbf t_t^o$.

Trajectory-level supervision alone does not guarantee physically plausible motion. We therefore further constrain the hand and object dynamics individually, as well as their physical coupling during interaction. Let $\hat{\mathbf x}_{1:T}$ denote the ground-truth trajectory. We formulate the reconstruction objective as
\begin{equation}
\mathbf x_{1:T}^{*}
=
\arg\min_{\mathbf x_{1:T}\in\mathcal X}
\left[
\mathcal C_t
\left(
\mathbf x_{1:T},
\hat{\mathbf x}_{1:T}
\right)
+
\lambda_d
\mathcal C_d
\left(
\mathbf x_{1:T},
\hat{\mathbf x}_{1:T}
\right)
+
\lambda_a
\mathcal C_a
\left(
\mathbf x_{1:T}^h,
\mathbf x_{1:T}^o
\right)
\right],
\label{eq:physics_formulation}
\end{equation}
where $\mathcal X$ denotes the admissible trajectory space. The three constraints capture complementary aspects of reconstruction: $\mathcal C_t$ enforces trajectory alignment, $\mathcal C_d$ independently constrains the articulated-hand and rigid-object dynamics, and $\mathcal C_a$ constrains their coupled dynamics during interaction. The weights $\lambda_d$ and $\lambda_a$ balance the corresponding physical constraints. Eq.~\ref{eq:physics_formulation} only provides a conceptual decomposition of constraints; in practice, they are instantiated as training objectives for the model.

\subsection{Physics-aware HOI Reconstruction}
\label{sec:framework}

Figure \ref{fig:method} shows the overall framework of \method{}, consisting of frozen reconstruction estimators and a learnable trajectory refinement network we design. Specifically, a frozen hand estimator $\mathcal E^h$ \citep{potamias2024wilor} takes the frame $I_t$ as input and predicts hand parameters $\mathbf y^h_t$, while a frozen object estimator $\mathcal E^o$ \citep{wen2024foundationpose} takes $I_t$ and the object template $\mathcal M^o$ as input and predicts the object parameters $\mathbf y^o_t$, resulting an initial estimate of the hand and object trajectory $\mathbf y_{1:T}=\left(\mathbf y^h_{1:T},\mathbf y^o_{1:T}\right)$. Because these estimations rely on framewise visual evidence, their predictions can suffer from temporal and physical inconsistency, particularly under monocular ambiguity and occlusion. We refine the initial estimates \citep{zhang2025diffusion,zhang2025hawor,ismayilzada2026padhand} by modeling visual-geometric evidence and physical characteristics through an iterative diffusion process.

\subsubsection{Refinement Network}
\label{sec:refinement}
We formulate the refinement within a diffusion process \citep{ho2020denoising,peebles2023scalable} that enables progressive correction of temporal incoherence through iterative denoising. Specifically, given the initial reconstruction $\mathbf y_{1:T}$, randomly-sampled diffusion steps $n$ and other inputs, the refinement network $f_{\Theta}$ estimates the temporally consistent trajectory by
\begin{equation}
\mathbf x^{(n)}_{1:T}
=
f_{\Theta}
\left(
\mathbf y_{1:T};
n,
I_{1:T},
\mathcal M^o
\right).
\label{eq:denoising_function}
\end{equation}
$f_{\Theta}$ contains a forward noise-injection process and a learnable denoising process, which progressively refines the initial  trajectory by visual-evidence grounding with 3D geometry and hand-object dependency modeling. During training, we first sample a diffusion step $n$ and construct the corresponding noisy trajectory $\mathbf z_{n,1:T}$ through the forward noise-injection process:
\begin{equation}
\mathbf z_{n,1:T}
=
\hat{\mathbf x}_{1:T}
+
\eta_n^2
\left(
\mathbf y_{1:T}
-
\hat{\mathbf x}_{1:T}
\right)
+
\eta_n\kappa\boldsymbol\epsilon,
\label{eq:forward_diffusion}
\end{equation}
where $\eta_n$ controls the diffusion level, $\kappa$ determines the stochastic noise magnitude, and $\boldsymbol\epsilon\sim\mathcal N(\mathbf 0,\mathbf I)$ denotes Gaussian noise. Similar to $\mathbf y_{1:T}$, we represent the noisy trajectory as $\mathbf z_{n,1:T}=\{(\mathbf z^h_{n,t},\mathbf z^o_{n,t})\}_{t=1}^{T}$, comprising the noisy hand trajectory $\mathbf z^h_{n,1:T}$ and the noisy object trajectory $\mathbf z^o_{n,1:T}$. The denoising process is performed through the following modules:

\textbf{Geometric encoding.}
We perform geometric encoding in mesh space to expose the explicit 3D structure hidden in the compact state parameters, enabling the denoising process to reason about hand-object spatial relationships and local surface geometry. The intermediate noisy states $\mathbf z_{n,1:T}$ are first converted into meshes by applying MANO layers to hand parameters and the rigid transformation to object parameters, which are subsequently encoded into latent features through separate geometry encoders. We further employ temporal self-attention layers to aggregate temporal information across the frame sequence, which produces the geometric feature sequences $\mathbf h^{\mathrm{geo}}_{n,1:T}$ and $\mathbf o^{\mathrm{geo}}_{n,1:T}$.

\textbf{Visual grounding.}
The geometric features capture explicit 3D structure but lack appearance information from the observed video. We therefore ground them with spatially corresponding visual evidence. Specifically, a frozen visual encoder~\citep{oquab2023dinov2} extracts a patch-level feature map $\mathbf F_t$ from each frame $I_t$. To establish spatial correspondence, we derive 3D geometric anchors from the initial hand and object estimates $\mathbf y^h_t$ and $\mathbf y^o_t$, using MANO joints for the hand and pose-transformed canonical keypoints for the object. These anchors are projected onto the image to retrieve and aggregate their corresponding visual features $\mathbf \zeta_t^h$ and $\mathbf \zeta_t^o$:
\begin{equation}
\mathbf \zeta_t^m
=
\frac{1}{|\mathcal Q_t^m|}
\sum_{\mathbf p\in\mathcal Q_t^m}
\mathbf F_t\!\left(\Pi_t(\mathbf p)\right),
\qquad
m\in\{h,o\},
\label{eq:proj_fusion}
\end{equation}
where $\mathcal Q_t^h$ and $\mathcal Q_t^o$ denote the sets of visible hand and object anchors, respectively, and $\Pi_t$ denotes the camera projection function. The resulting geometry-aligned visual features are fused with their corresponding geometric features through residual projections:
$\mathbf h^{\mathrm{vis}}_{n,t}
=
\mathbf h^{\mathrm{geo}}_{n,t}
+
\phi^h(\mathbf \zeta_t^h)$
and
$\mathbf o^{\mathrm{vis}}_{n,t}
=
\mathbf o^{\mathrm{geo}}_{n,t}
+
\phi^o(\mathbf \zeta_t^o)$,
where $\phi^h$ and $\phi^o$ are learnable projection layers. This produces the visually grounded feature sequences $\mathbf h^{\mathrm{vis}}_{n,1:T}$ and $\mathbf o^{\mathrm{vis}}_{n,1:T}$.

\textbf{Interaction modeling.}
Although the hand and object features are visually grounded, they are still encoded separately, which does not explicitly capture their mutual dependencies. We therefore concatenate $\mathbf h^{\mathrm{vis}}_{n,1:T}$ and $\mathbf o^{\mathrm{vis}}_{n,1:T}$ and jointly process them with spatial-temporal self-attention layers \citep{dosovitskiy2020image} to model HOI across time. The output tokens are then separated into the interaction-aware hand and object features according to the original token positions, with the first $T$ tokens forming $\tilde{\mathbf h}_{n,1:T}$ and the remaining $T$ tokens forming $\tilde{\mathbf o}_{n,1:T}$. Separate prediction heads regress residual updates to the initial hand and object states as
$\mathbf x^{h,(n)}_t=\mathbf y^h_t+g^h(\tilde{\mathbf h}_{n,t})$
and
$\mathbf x^{o,(n)}_t=\mathbf y^o_t+g^o(\tilde{\mathbf o}_{n,t})$,
where $g^h$ and $g^o$ are corresponding MLP projectors. We concatenate them as $\mathbf x^{(n)}_t=[\mathbf x^{h,(n)}_t,\mathbf x^{o,(n)}_t]$ and collect over frames to obtain $\mathbf x^{(n)}_{1:T}=\{\mathbf x^{(n)}_t\}^T_{t=1}$ as the output of $f_{\Theta}$.

\textbf{Diffusion objective.}
We instantiate the trajectory-level constraint $\mathcal C_t$ in Eq.~\ref{eq:physics_formulation} with the diffusion objective $\mathcal L_{\mathrm{diff}}$. Specifically, we train the diffusion-based refinement network $f_{\Theta}$ to recover the ground-truth trajectory from noisy trajectories sampled at different diffusion levels. We align the estimated trajectory $\mathbf x^{(n)}_{1:T}$ with the ground-truth $\hat{\mathbf x}_{1:T}$ by
\begin{equation}
\mathcal L_{\mathrm{diff}}
=
\mathbb E_{n,\boldsymbol\epsilon}
\left[
d_x
\left(
\hat{\mathbf x}_{1:T},
\mathbf x^{(n)}_{1:T}
\right)
\right],
\label{eq:ldiff}
\end{equation}
where $d_x$ measures the trajectory discrepancy in rotation and translation spaces. This objective enables $f_{\Theta}$ to recover accurate HOI trajectories through joint visual-geometric representation learning.

\subsubsection{Hand and Object Dynamics}
\label{sec:dynamics}

Although the diffusion objective enables trajectory reconstruction, it does not explicitly account for the underlying dynamics that produce the hand and object motion, which may lead to physically implausible behaviors such as abnormal accelerations or abrupt rotations. To promote physical consistency beyond visual plausibility, we derive the hand and object dynamics from their trajectories and align the resulting physical quantities between the reconstructed and ground-truth motions.

\textbf{Hand dynamics.}
Following previous work \citep{zhang2024physpt,ismayilzada2026padhand}, we regard the human hand as an articulated body and derive the generalized force required to produce its motion. The 16 hand rotations $\boldsymbol\rho^h_t$ are converted into Euler-ZXY angles $\boldsymbol\phi^h_t\in\mathbb R^{48}$, corresponding to the global orientation and $15$ articulated MANO joints. Together with the global translation $\mathbf t^h_t$, we define the generalized coordinates as
$\mathbf q^h_t=[(\mathbf t^h_t)^\top,(\boldsymbol\phi^h_t)^\top]^\top\in\mathbb R^{51}$.
We use temporal finite differences between adjacent frames to estimate the generalized velocity $\dot{\mathbf q}^h_t$ and acceleration $\ddot{\mathbf q}^h_t$. Given these kinematic quantities, inverse dynamics is used to recover the resultant generalized force:
\begin{equation}
\boldsymbol\tau^h_t
=
M^h(\mathbf q^h_t)\ddot{\mathbf q}^h_t
+
C^h(\mathbf q^h_t,\dot{\mathbf q}^h_t)\dot{\mathbf q}^h_t
+
\mathbf g^h_{\mathrm{grav}}(\mathbf q^h_t),
\label{eq:id}
\end{equation}
where $M^h$, $C^h$, and $\mathbf g^h_{\mathrm{grav}}$ denote the hand mass matrix, Coriolis and centrifugal terms, and gravitational term, respectively. Applying Eq.~\ref{eq:id} over the frame sequence yields the generalized-force trajectory $\{\boldsymbol\tau^h_t\}_{t=1}^{T}$, which reveals the dynamic requirements for driving hand movement.

\textbf{Object dynamics.}
We introduce the Newton--Euler formalism~\citep{luh1980line,khalil2011dynamic, hu2022physicalinteraction,lutter2023differentiable}, a fundamental framework for inverse dynamics in robotics and rigid-body control, into HOI reconstruction to relate object motion to its underlying physical requirements. Specifically, as discussed in Sect.~\ref{sec:formulation}, we model the object as a rigid body with pose $\mathbf x_t^o=(\mathbf R_t^o,\mathbf t_t^o)$. We compute its center-of-mass linear acceleration $\mathbf a^o_t$ from the translation trajectory $\{\mathbf t^o_t\}_{t=1}^{T}$, and derive its angular velocity $\boldsymbol{\omega}^o_t$ and angular acceleration $\dot{\boldsymbol{\omega}}^o_t$ from the rotation trajectory $\{\mathbf R^o_t\}_{t=1}^{T}$ using temporal finite differences between adjacent frames. The physical effort required to produce this motion is thus represented by its object wrench:
\begin{equation}
\mathbf w^o_t
=
\begin{bmatrix}
\mathbf f^o_t\\
\boldsymbol{\tau}^o_t
\end{bmatrix}
=
\begin{bmatrix}
m^o(\mathbf a^o_t-\mathbf g)\\
\mathbf I^o_t\dot{\boldsymbol{\omega}}^o_t
+
\boldsymbol{\omega}^o_t\times
(\mathbf I^o_t\boldsymbol{\omega}^o_t)
\end{bmatrix},
\label{eq:wrench}
\end{equation}
where $\mathbf f^o_t$ and $\boldsymbol{\tau}^o_t$ denote the non-gravitational external force and torque required to drive the translational and rotational motion of the object, respectively. $\mathbf g$ is the gravitational acceleration, $\mathbf I^o_t$ is the inertia tensor in the world frame, and $m^o$ denotes the object mass. We use the object mass provided by the dataset when available; otherwise, we employ a large language model \citep{openai2026gpt56} to estimate it. This formulation maps the object-pose trajectory to its wrench profile $\{\mathbf w^o_t\}_{t=1}^{T}$, providing an explicit dynamics-based characterization of object motion. Additional theoretical and implementation details are provided in the supplementary material.

\textbf{Dynamics objective.}
The hand generalized forces and object wrenches derived above provide a dynamics-level description of the motion. We instantiate the dynamics constraint $\mathcal C_d$ in Eq.~\ref{eq:physics_formulation} by aligning the mechanical quantities implied by the reconstructed trajectory with those derived from the ground-truth trajectory. For frame $t$, we compute the hand generalized force and object wrench $(\boldsymbol\tau_t^{h,(n)}, \mathbf w_t^{o,(n)})$ at diffusion step $n$ from the predicted trajectory, and $(\hat{\boldsymbol\tau}_t^{h}, \hat{\mathbf w}_t^{o})$ from the ground-truth trajectory. The resulting dynamics objective is
\begin{equation}
\mathcal L_{\mathrm{dyn}}
=
\sum_{t=1}^T
\left\|
\boldsymbol\tau_t^{h,(n)}-\hat{\boldsymbol\tau}_t^{h}
\right\|_2^2
+
\sum_{t=1}^T
\left\|
\mathbf w_t^{o,(n)}-\hat{\mathbf w}_t^{o}
\right\|_2^2.
\label{eq:ldyn}
\end{equation}
This objective complements the diffusion supervision by enforcing consistency in the mechanical quantities underlying the motion, thereby discouraging dynamically inconsistent trajectories.

\subsubsection{Coupled HOI Dynamics}
\label{sec:actuation}

\begin{figure}[t]
    \centering
    \includegraphics[width=0.6\linewidth]{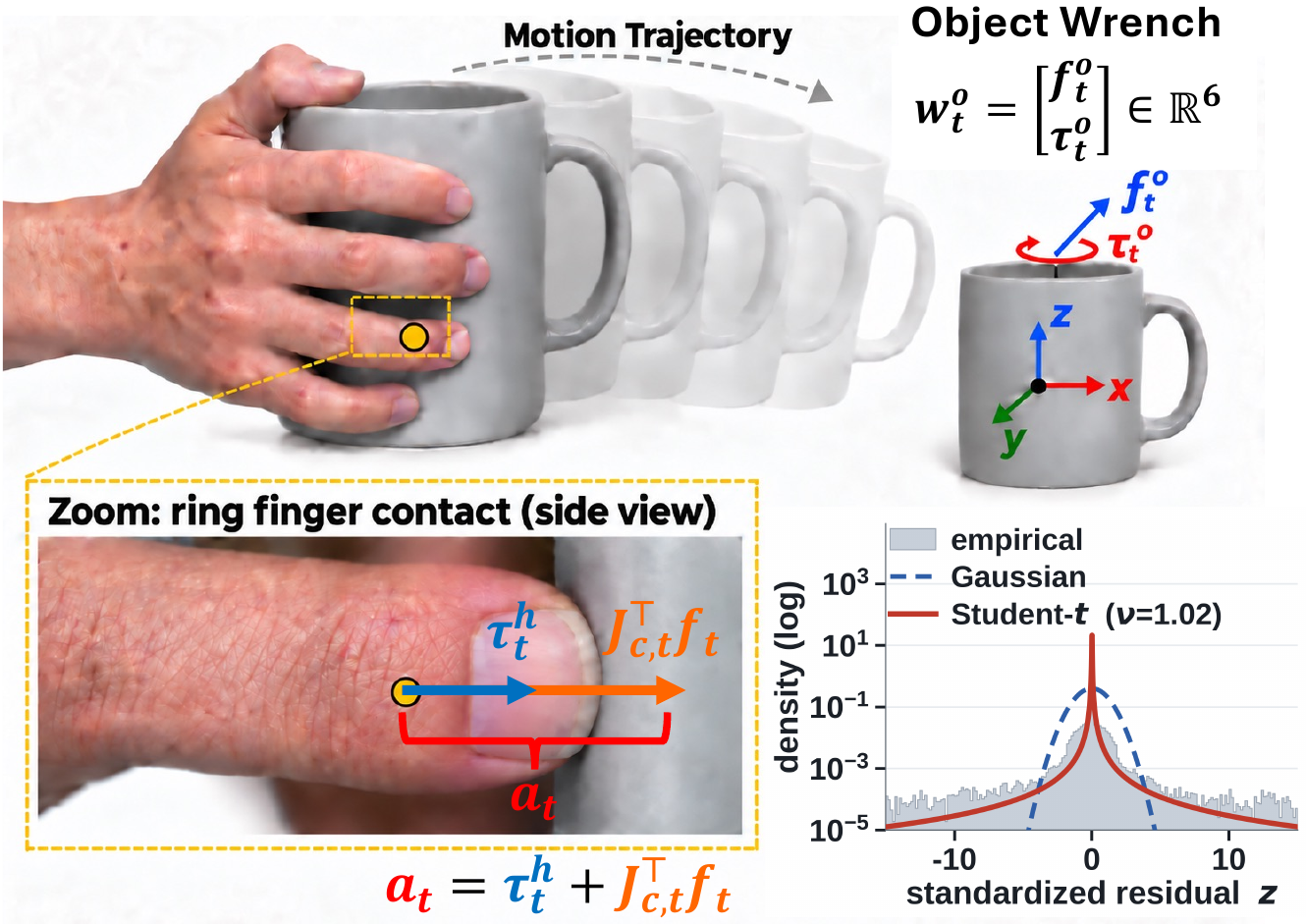}
  \caption{Coupled HOI dynamics. The bottom-right plot shows the empirical distribution of standardized active actuation $z=(a_{t,d}-\mu_d)/\sigma_d$ for an example DoF. The logarithmic density axis highlights the heavy-tailed behavior, which is empirically better captured by the Student-$t$ distribution than the Gaussian distribution.}
  \label{fig:force}
\end{figure}

Even if individually plausible, hand and object dynamics may still be mutually inconsistent, since physically valid HOI requires the same interaction to explain both motions~\citep{hu2022physicalinteraction,nakajima2022prediction}. Active hand actuation provides such a joint physical description by accounting for both the generalized force required to produce hand motion and the interaction force exerted to drive object motion. Therefore, we use it to couple dynamics and regularize it with a probabilistic prior~\citep{tanghe2019probabilistic} to encourage physically plausible joint motion.

To recover active hand actuation, we first infer the contact forces exerted by the hand to drive the object motion. Since we focus on in-air manipulation, where the object is primarily driven by hand contact, we do not explicitly model external supports such as table contact. We approximate distributed contact using $K=6$ anchors on the hand mesh, including the five fingertips and palm center. Given the contact geometry at frame $t$, we infer $\mathbf f_t\in\mathbb R^{3K}$ whose combined effect explains the object wrench $\mathbf w_t^o$. These contact forces are mapped to the hand generalized coordinates through $J_{c,t}^{\top}\mathbf f_t$ and combined with the generalized force $\boldsymbol\tau_t^h$ required by the hand motion, yielding the active hand actuation $\mathbf a_t\in\mathbb R^{51}$:
\begin{equation}
\mathbf a_t
=
\boldsymbol\tau_t^{h}
+
J_{c,t}^{\top}\mathbf f_t,
\label{eq:actuation}
\end{equation}
which jointly accounts for the physical effort underlying the hand and object motion, thereby coupling their dynamics. More details of contact-force inference are provided in the supplementary material.

We assume that hand forces during everyday object manipulation typically remain within a characteristic range, and thus active actuation should follow a structured distribution rather than vary arbitrarily. We estimate this distribution from framewise actuation computed on ground-truth training trajectories. As shown in Figure~\ref{fig:force}, most samples concentrate within a narrow range, while larger values occasionally occur during dynamic interactions such as lifting or releasing objects, yielding a heavy-tailed distribution well captured by a Student-$t$ distribution empirically. Accordingly, we instantiate the coupled-dynamics constraint $\mathcal C_a$ in Eq.~\ref{eq:physics_formulation} by fitting a separate Student-$t$ distribution to each DoF and using its negative log-likelihood to penalize unlikely actuation. For coordinate $d\in\{1,\ldots,51\}$ with fitted parameters $(\mu_d,\sigma_d,\nu_d)$, the active-actuation objective is
\begin{equation}
\mathcal L_{\mathrm{act}}
=
\sum_{t,d}
\alpha_t \hat{c}_t
\left[
\log\sigma_d
+
\frac{\nu_d+1}{2}
\log
\left(
1+
\frac{(a_{t,d}-\mu_d)^2}
{\nu_d\sigma_d^2}
\right)
\right],
\label{eq:lact}
\end{equation}
where $a_{t,d}$ is the $d$-th DoF of $\mathbf a_t$, $\hat{c}_t\in\{0,1\}$ indicates ground-truth hand-object contact, and $\alpha_t$ emphasizes dynamically active frames. Since $\mathbf a_t$ jointly depends on hand motion, object motion, and contact geometry, this objective provides a learned prior over their coupled dynamics while remaining tolerant to legitimate high-actuation events.

\subsection{Training Objective and Inference}
\label{sec:loss}

We jointly optimize three objectives to train the model in an end-to-end fashion, which is defined as:
\begin{equation}
\mathcal L
=
\mathcal L_{\mathrm{diff}}
+
\lambda_{\mathrm{d}}\mathcal L_{\mathrm{dyn}}
+
\lambda_{\mathrm{a}}\mathcal L_{\mathrm{act}},
\label{eq:total_loss}
\end{equation}
where $\mathcal L_{\mathrm{diff}}$ supervises trajectory reconstruction, $\mathcal L_{\mathrm{dyn}}$ constrains the individual hand and object dynamics, and $\mathcal L_{\mathrm{act}}$ further enforces their coupled mechanics through contact. Together, these objectives realize our formulated problem in Eq.~\ref{eq:physics_formulation}, which encourages reconstructed trajectories that are both geometrically accurate and physically consistent. The physical quantities are only calculated during training, thus no additional computational cost is introduced at inference.

\section{Experiments}
\label{sec:experiments}

\subsection{Experimental Setting}

\textbf{Datasets and protocols.}
We conduct experiments on an egocentric video dataset HOT3D~\citep{banerjee2025hot3d}, and two allocentric video datasets HO3D-v2~\citep{hampali2020honnotate} and DexYCB~\citep{chao2021dexycb}. For HOT3D, we follow the Aria video split \citep{ye2026whole}, using 1,311 training and 147 evaluation clips, each with 150 frames. For HO3D-v2, we use the official split with 55 training sequences and 13 evaluation sequences (11,524 frames). For DexYCB, we follow the official protocol and treat each camera view as a clip, yielding 6,400/320/1,280 clips for training/validation/testing.

\textbf{Evaluation metrics.}
For hand reconstruction, HOT3D reports W-MPJPE (W; cm), WA-MPJPE (WA; cm), PA-MPJPE (PA; mm), RTE (\%), and ACCEL (mm/frame$^2$), where W and WA align trajectories using the first two frames and the full sequence, respectively. HO3D-v2 reports PA-MPJPE for both joints and vertices (PA/PA-V; mm), their corresponding AUC/AUC-V (in $[0,1]$), F@5/F@15, and ACCEL (mm/frame$^2$). DexYCB reports wrist-relative MPJPE, PA-MPJPE (PA; mm), AUC (in $[0,1]$), and ACCEL (mm/frame$^2$). Object pose is evaluated using ADD(-S) AUC (\%) at thresholds of 0.1/0.3\,m, 0.1d recall (\%), ACCEL (mm/frame$^2$) and $5^\circ$/5\,cm success (\%). More evaluation and implementation details are provided in the supplementary material.

\begin{table}[t]
\centering
\caption{Comparison on HOT3D, HO3D-v2, and DexYCB datasets. \textbf{Bold} / \underline{underline} denote the best / second-best reconstruction methods. $\dagger$ denotes our reproduction, and unmarked values are their paper-reported. “--” denotes metrics that are either inapplicable to the method or unavailable due to missing reports and unreleased code.}
\label{tab:comparison}
\begin{subtable}{\linewidth}
\centering
\caption{Hand-object reconstruction on HOT3D dataset.}
\label{tab:hot3d}
\tablesize
\setlength{\tabcolsep}{0pt}

\begin{tabular*}{\linewidth}{@{\extracolsep{\fill}}lcccccccccccc@{}}
\toprule
&\multicolumn{5}{c}{Hand}&\multicolumn{7}{c}{Object}\\
\cmidrule(lr){2-6}\cmidrule(lr){7-13}
Method&W$\downarrow$&WA$\downarrow$&PA$\downarrow$&ACCEL$\downarrow$&RTE$\downarrow$
&ADD@.1$\uparrow$&ADD@.3$\uparrow$&ADD-S@.1$\uparrow$&ADD-S@.3$\uparrow$&0.1d$\uparrow$&ACCEL$\downarrow$&$5^\circ$5cm$\uparrow$\\
\midrule
\multicolumn{13}{@{}l}{\itshape Input estimates ($\mathbf y$)}\\
WiLoR$\repro$&11.74&3.70&6.52&16.22&75.16&--&--&--&--&--&--&--\\
FoundationPose$\repro$&--&--&--&--&--&12.8&38.9&25.3&52.8&6.5&70.15&6.7\\
\midrule
\multicolumn{13}{@{}l}{\itshape Reconstruction methods}\\
HaWoR$\repro$&11.73&3.92&8.38&\underline{12.90}&\underline{73.69}&--&--&--&--&--&--&--\\
WHOLE&\underline{10.41}&\underline{3.26}&\underline{6.67}&--&--&--&\underline{51.1}&--&\underline{69.9}&--&--&--\\
\textbf{\method}&
\textbf{5.26}&
\textbf{1.83}&
\textbf{4.15}&
\textbf{1.68}&
\textbf{73.49}&
\textbf{25.8}&
\textbf{65.4}&
\textbf{48.2}&
\textbf{78.4}&
\textbf{7.1}&
\textbf{5.57}&
\textbf{9.8}\\
\bottomrule
\end{tabular*}
\end{subtable}

\vspace{2mm}

\begin{subtable}[t]{0.55\linewidth}
\centering
\caption{Hand reconstruction on HO3D-v2 dataset.}
\label{tab:ho3d}
\tablesize
\setlength{\tabcolsep}{2.5pt}
\begin{tabular}{@{}lccccccc@{}}
\toprule
Method&PA$\downarrow$&AUC$\uparrow$&PA-V$\downarrow$&AUC-V$\uparrow$&F@5$\uparrow$&F@15$\uparrow$&ACCEL$\downarrow$\\
\midrule
\multicolumn{8}{@{}l}{\itshape Input estimates ($\mathbf y$)}\\
WiLoR$\repro$&7.67&0.847&7.66&0.846&0.647&0.984&4.12\\
\midrule
\multicolumn{8}{@{}l}{\itshape Reconstruction methods}\\
AMVUR&8.30&0.835&8.20&0.836&0.608&0.965&--\\
HaMeR&7.70&0.846&7.90&0.841&0.635&0.980&--\\
Hamba&\underline{7.50}&\underline{0.850}&\underline{7.70}&\underline{0.846}&\underline{0.648}&\underline{0.982}&--\\
\textbf{\method}&
\textbf{7.45}&
\textbf{0.851}&
\textbf{7.49}&
\textbf{0.850}&
\textbf{0.656}&
\textbf{0.985}&
\textbf{2.77}\\
\bottomrule
\end{tabular}
\end{subtable}
\hfill
\begin{subtable}[t]{0.4\linewidth}
\centering
\caption{Hand reconstruction on DexYCB dataset.}
\label{tab:dexycb}
\tablesize
\setlength{\tabcolsep}{3pt}
\begin{tabular}{@{}lcccc@{}}
\toprule
Method&MPJPE$\downarrow$&PA$\downarrow$&AUC$\uparrow$&ACCEL$\downarrow$\\
\midrule
\multicolumn{5}{@{}l}{\itshape Input estimates ($\mathbf y$)}\\
WiLoR$\repro$&11.52&5.33&0.893&6.88\\
\midrule
\multicolumn{5}{@{}l}{\itshape Reconstruction methods}\\
Deformer&13.64&5.22&--&6.77\\
HaWoR$\repro$&
11.09&
5.32&
\underline{0.894}&
3.88\\
PAD-Hand&
\underline{10.56}&
\textbf{4.63}&
--&
\underline{3.34}\\
\textbf{\method}&
\textbf{9.51}&
\underline{4.68}&
\textbf{0.906}&
\textbf{3.27}\\
\bottomrule
\end{tabular}
\end{subtable}
\end{table}

\subsection{Experimental Results Analysis}
\textbf{Model Comparison.}
We compare \method with existing reconstruction methods, including WiLoR~\citep{potamias2024wilor}, FoundationPose~\citep{wen2024foundationpose}, HaWoR~\citep{zhang2025hawor}, WHOLE~\citep{ye2026whole}, AMVUR~\citep{jiang2023probabilistic}, HaMeR~\citep{pavlakos2024hamer}, Hamba~\citep{dong2024hamba}, Deformer~\citep{fu2023deformer}, and PAD-Hand~\citep{ismayilzada2026padhand}. Table~\ref{tab:comparison} shows that \method achieves the best performance on most metrics while consistently improving the input hand and object estimates. The gains are particularly large on the egocentric HOT3D dataset, indicating that our model remains effective under moving viewpoints. Importantly, modeling object dynamics and coupling them with hand dynamics improves not only object reconstruction but also hand motion. On the DexYCB dataset, \method outperforms PAD-Hand, which explicitly models hand dynamics, suggesting that object dynamics provide complementary constraints on hand reconstruction through physical interaction. This benefit is further reflected in the reduction in acceleration error (ACCEL) across all datasets for both hand and object, indicating that our proposed physical constraints improve not only trajectory accuracy but also the dynamical consistency of hand-object motion.

\textbf{Module Ablation.}
As shown in Table~\ref{tab:ablate_modules}, for hand reconstruction, removing visual grounding causes the largest degradation on egocentric HOT3D, indicating that geometry-aligned visual evidence is particularly important for correcting trajectories in a moving camera. In contrast, the interaction module brings a minor effect, because dense hand-object co-motion can already be captured by temporal refinement. On fixed-view HO3D, removing interaction causes the largest degradation, as persistent occlusion and partial visibility make explicit hand-object motion coupling more important, while geometric encoding is less critical under a stable camera. For object reconstruction on HOT3D, interaction is the most influential module, showing that hand motion provides strong constraints for recovering object pose. Geometric encoding has a smaller effect, because the rigid object structure is already well preserved, leaving the other two modules as main cues for resolving its pose changes.

\begin{table}[t]
\centering
\caption{Module ablation. \textbf{Bold} / \underline{underline} denote the best / second-best configurations.}
\label{tab:ablate_modules}

\begin{subtable}[t]{0.68\linewidth}
\centering
\caption{Ablations for hand-object reconstruction on the HOT3D dataset.}
\tablesize
\setlength{\tabcolsep}{2.8pt}
\begin{tabular}{@{}lcccccccc@{}}
\toprule
&\multicolumn{4}{c}{Hand}&\multicolumn{4}{c}{Object}\\
\cmidrule(lr){2-5}\cmidrule(l){6-9}
Model&W$\downarrow$&WA$\downarrow$&PA$\downarrow$&ACCEL$\downarrow$
&ADD@0.1$\uparrow$&ADD-S@0.1$\uparrow$&0.1d$\uparrow$&$5^\circ$5cm$\uparrow$\\
\midrule
w/o Geometric &5.83&1.92&4.33&\underline{1.71}&\underline{24.4}&\underline{46.0}&\textbf{7.1}&\underline{9.5}\\
w/o Grounding&6.65&2.40&4.45&1.74&21.5&43.0&5.3&8.1\\
w/o Interaction&\underline{5.72}&\underline{1.90}&\underline{4.24}&1.73&21.2&40.0&\underline{5.5}&8.0\\
\midrule
\textbf{Full model}&\textbf{5.26}&\textbf{1.83}&\textbf{4.15}&\textbf{1.68}&\textbf{25.8}&\textbf{48.2}&\textbf{7.1}&\textbf{9.8}\\
\bottomrule
\end{tabular}
\end{subtable}
\hfill
\begin{subtable}[t]{0.30\linewidth}
\centering
\caption{Ablations for hand reconstruction on the HO3D dataset.}
\tablesize
\setlength{\tabcolsep}{3.0pt}
\begin{tabular}{@{}lccc@{}}
\toprule
Model&PA$\downarrow$&PA-V$\downarrow$&F@5$\uparrow$\\
\midrule
w/o Geometric&\underline{7.52}&\underline{7.51}&\underline{0.655}\\
w/o Grounding&7.55&\underline{7.51}&0.654\\
w/o Interaction&7.59&7.52&0.654\\
\midrule
\textbf{Full model}&\textbf{7.45}&\textbf{7.49}&\textbf{0.656}\\
\bottomrule
\end{tabular}
\end{subtable}
\end{table}

\begin{table}[t]
\centering
\caption{Loss ablation. \textbf{Bold} / \underline{underline} denote the best / second-best configurations.}
\label{tab:loss_ablation}
\begin{subtable}[t]{0.645\linewidth}
\centering
\caption{Ablations for hand-object reconstruction on the HOT3D dataset.}
\tablesize
\setlength{\tabcolsep}{2pt}
\begin{tabular}{@{}ccc cccc cccc@{}}
\toprule
\multicolumn{3}{c}{Training objective} & \multicolumn{4}{c}{Hand} & \multicolumn{4}{c}{Object} \\
\cmidrule(lr){1-3} \cmidrule(lr){4-7} \cmidrule(l){8-11}
$\mathcal{L}_{\text{diff}}$ & $\mathcal{L}_{\text{dyn}}$ & $\mathcal{L}_{\text{act}}$ & W$\downarrow$ & WA$\downarrow$ & PA$\downarrow$ & ACCEL$\downarrow$ & ADD@0.1$\uparrow$ & ADD-S@0.1$\uparrow$ & 0.1d$\uparrow$ & $5^\circ$5cm$\uparrow$ \\
\midrule
\checkmark & & & 8.27 & 2.82 & 5.16 & 6.10 & 17.8 & 33.6 & 4.7 & 7.4 \\
\checkmark & & \checkmark & 7.77 & 2.31 & 5.06 & 5.01 & 20.0 & 37.9 & 5.3 & 8.5 \\
\checkmark & \checkmark & & \underline{6.12} & \underline{1.91} & \underline{4.22} & \underline{2.78} & \underline{24.9} & \underline{47.4} & \underline{6.7} & \textbf{9.8} \\
\checkmark & \checkmark & \checkmark & \textbf{5.26} & \textbf{1.83} & \textbf{4.15} & \textbf{1.68} & \textbf{25.8} & \textbf{48.2} & \textbf{7.1} & \textbf{9.8} \\
\bottomrule
\end{tabular}
\end{subtable}
\hfill
\begin{subtable}[t]{0.33\linewidth}
\centering
\caption{Ablations for hand on HO3D.}
\tablesize
\setlength{\tabcolsep}{1.6pt}
\begin{tabular}{@{}ccc ccc@{}}
\toprule
\multicolumn{3}{c}{Training objective} & \multicolumn{3}{c}{Hand} \\
\cmidrule(lr){1-3} \cmidrule(l){4-6}
$\mathcal{L}_{\text{diff}}$ & $\mathcal{L}_{\text{dyn}}$ & $\mathcal{L}_{\text{act}}$ & PA$\downarrow$ & PA-V$\downarrow$ & F@5$\uparrow$ \\
\midrule
\checkmark & & & 7.55 & 7.59 & 0.650 \\
\checkmark & & \checkmark & 7.48 & \underline{7.50} & 0.651 \\
\checkmark & \checkmark & & \underline{7.46} & \underline{7.50} & 0.651 \\
\checkmark & \checkmark & \checkmark & \textbf{7.45} & \textbf{7.49} & \textbf{0.656} \\
\bottomrule
\end{tabular}
\end{subtable}
\end{table}

\begin{figure}[t]
    \centering
    \includegraphics[width=0.85\linewidth]{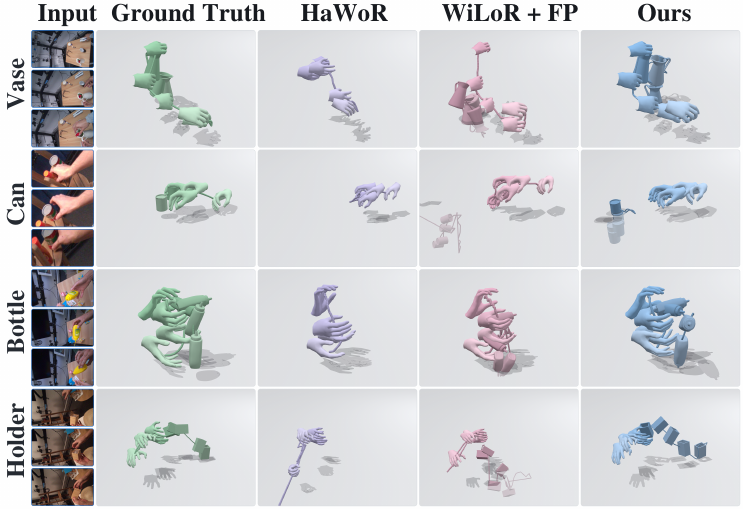}
  \caption{Visualization for reconstruction.}
  \label{fig:vis}
\end{figure}

\begin{table}[t]
\centering
\caption{Ablation of different distribution choices for modeling active hand actuation on HOT3D dataset.}
\label{tab:ablation_distribution}
\tablesize
\begin{tabular}{lccc}
\toprule
Distribution & NLL $\downarrow$ & PA $\downarrow$ & ADD@.1 $\uparrow$ \\
\midrule
w/o $\mathcal{L}_{\text{act}}$ & -- & 4.22 & 24.9 \\
\midrule
Gaussian            & \textminus0.39 & 4.54 & 24.2 \\
Laplace             & \textminus0.11 & 4.17 & 24.9 \\
Cauchy ($\nu{=}1$)  & \underline{\textminus1.45} & 4.20 & 24.7 \\
\midrule
\multicolumn{4}{l}{\textit{Student-$t$ distributions}} \\
\quad Fixed $\nu{=}10$        & \textminus0.70 & 4.17 & \underline{25.6} \\
\quad Fixed $\nu{=}5$         & \textminus0.96 & \underline{4.16} & 25.2 \\
\quad Fixed $\nu{=}2$         & \textminus1.15 & \underline{4.16} & 25.3 \\
\quad Fitted $\nu_d$          & \textbf{\textminus1.59} & \textbf{4.15} & \textbf{25.8} \\
\bottomrule
\end{tabular}
\end{table}

\textbf{Loss Ablation.}
As shown in Table~\ref{tab:loss_ablation}, adding $\mathcal L_{\rm dyn}$ (2nd vs. 4th rows) substantially improves both hand and object reconstruction on HOT3D, confirming the benefit of explicitly constraining their dynamics. Comparing the last two rows further shows that $\mathcal L_{\rm act}$ provides a smaller but consistent additional gain. We attribute this limited improvement partly to the large variability of video content and to errors in the forces estimated by the Euler-Lagrange and Newton-Euler equations, which may propagate into the inferred actuation distribution. More accurate force estimation and actuation modeling remain promising directions for future work. On fixed-view HO3D, both physical losses yield modest gains, suggesting that dynamics supervision is particularly beneficial for egocentric reconstruction, where camera motion and trajectory ambiguity make purely visual constraints insufficient.

\textbf{Actuation Distribution.}
As shown in Table~\ref{tab:ablation_distribution}, we compare actuation priors using the average NLL per valid training coordinate and PA/ADD@0.1 reconstruction metrics. We can observe that an inappropriate prior can degrade reconstruction: the Gaussian prior underperforms removing $\mathcal L_{\rm act}$, supporting the heavy-tailed nature of the inferred actuation. Fitting $\nu_d$ separately for each DoF consistently outperforms fixed-$\nu$ Student-$t$ priors and achieves the best overall performance. Notably, better likelihood fit does not necessarily imply better reconstruction: although $\nu=2$ yields lower NLL than $\nu=10$, it performs worse on object reconstruction, suggesting that likelihood fit alone does not determine the effectiveness of the prior as a reconstruction constraint.

\subsection{Downstream Applications}
\label{sec:applications}

Beyond reconstruction, we evaluate whether physics-aware hand trajectories from \method benefit downstream embodied tasks. For dexterous manipulation, we follow ViViDex~\citep{chen2025vividex} and retarget trajectories reconstructed on the DexYCB dataset for robot policy learning. For egocentric world modeling, we follow EgoHOI~\citep{li2026egohoi} and use trajectories reconstructed on the HOT3D dataset as hand-action conditions. In both tasks, only the input hand trajectories vary across reconstruction methods, while all other components remain fixed. We use reconstructed trajectories from models trained on the corresponding dataset without task-specific retraining.

\begin{figure}[t]
\centering
\captionsetup[sub]{justification=centering}
\begin{minipage}[b]{0.44\textwidth}
  \centering
  \includegraphics[width=\linewidth]{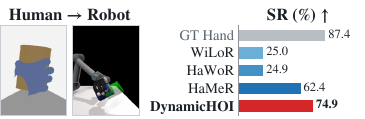}
  \subcaption{Dexterous manipulation.}
  \label{fig:downstream_dex}
\end{minipage}\hfill
\begin{minipage}[b]{0.54\textwidth}
  \centering
  \includegraphics[width=\linewidth]{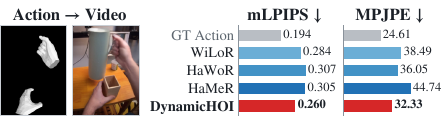}
  \subcaption{Egocentric video generation.}
  \label{fig:downstream_wm}
\end{minipage}
\caption{Downstream applications with different hand-trajectory sources.}
\label{fig:downstream}
\end{figure}

\textbf{Dexterous manipulation.}
We evaluate eight held-out DexYCB relocate tasks, training one PPO policy~\citep{schulman2017proximal} for each object and trajectory source and evaluating 100 episodes per task. As shown in Figure~\ref{fig:downstream_dex}, downstream success provides a task-level measure of reconstruction quality. Methods with comparable reconstruction accuracy, such as WiLoR and HaWoR (see Table~\ref{tab:dexycb}), also achieve similar success rates. In contrast, \method reduces MPJPE by 17.4\% over WiLoR, and this advantage is amplified in embodied learning, increasing the success rate from 25.0\% to 74.9\%. This suggests that improvements in physically coherent hand reconstruction can translate into substantially larger gains when the trajectories are used for robotic learning.

\textbf{Egocentric world modeling.}
We evaluate both action fidelity and generated hand appearance. mLPIPS measures perceptual similarity within the GT hand region, while MPJPE measures the kinematic accuracy of the generated hand. Figure~\ref{fig:downstream_wm} shows that \method performs best among reconstructed action sources on all metrics. Since EgoHOI conditions generation on framewise MANO renderings aligned with image pixels, more accurate trajectories provide better spatial guidance for where and how the hand should appear. Thus, improved reconstruction not only yields more faithful generated motion but also translates into better visual quality in the hand region.

\section{Conclusion}
\label{sec:conclusion}

We presented \method, a physics-aware HOI reconstruction framework that combines geometry-grounded trajectory refinement with coupled hand-object dynamics. We model articulated-hand and rigid-object dynamics and couple them through contact, with a learned probabilistic prior over active hand actuation. Experiments on egocentric and allocentric datasets demonstrate improved reconstruction, while downstream gains in dexterous manipulation and egocentric world modeling show the broader value of physically coherent HOI trajectories. Future work will explore more accurate force estimation, richer actuation priors, and unified object representations that generalize beyond known template meshes to unseen objects.

\bibliographystyle{plainnat}
\bibliography{references}

\end{document}